\documentclass[conference]{IEEEtran}
\usepackage{cite}
\usepackage{amsmath,amssymb,amsfonts}
\usepackage{algorithmic}
\usepackage{graphicx}
\usepackage{textcomp}
\usepackage{xcolor}
\usepackage{url}
\def\BibTeX{{\rm B\kern-.05em{\sc i\kern-.025em b}\kern-.08em
    T\kern-.1667em\lower.7ex\hbox{E}\kern-.125emX}}
\begin{document}

\title{TWIG: Towards pre-hoc Hyperparameter Optimisation and Cross-Graph Generalisation via Simulated KGE Models}
% {\footnotesize \textsuperscript{*}Note: Sub-titles are not captured in Xplore and
% should not be used}
% \thanks{Identify applicable funding agency here. If none, delete this.}
% }

\author{\IEEEauthorblockN{1\textsuperscript{st} Jeffrey Sardina}
\IEEEauthorblockA{\textit{School of Computer Science and Statistics} \\
\textit{Trinity College Dublin}\\
Dublin, Ireland \\
0000-0003-0654-2938}
\and
\IEEEauthorblockN{2\textsuperscript{nd} John D. Kelleher}
\IEEEauthorblockA{\textit{School of Computer Science and Statistics} \\
\textit{Trinity College Dublin}\\
Dublin, Ireland \\
0000-0001-6462-3248}
\and
\IEEEauthorblockN{3\textsuperscript{rd} Declan O'Sullivan}
\IEEEauthorblockA{\textit{School of Computer Science and Statistics} \\
\textit{Trinity College Dublin}\\
Dublin, Ireland \\
0000-0003-1090-3548}
}

\maketitle

\begin{abstract}
Knowledge Graphs (KGs) have become ever-more important for modelling biomedical information, as their intrinsic graph structure matches the structure of many biological interaction networks. Together with KGs, Knowledge Graph Embeddings (KGEs) have shown immense potential to learn biological data and predict new, in-band facts about the data the KG describes. However, recent literature has suggested several major deficits to KGEs: that they have an extremely short `receptive field' of data they use to make predictions and that their learning is guided by memorising graph structure, not learning latent semantics. Moreover, while several studies have suggested that graph structure and KGE model choice affect optimal hyperparameters, the exact relationship of hyperparameters to learning remains unknown and is instead solved using a computationally intensive hyperparameter search.

In this paper we introduce TWIG (Topologically-Weighted Intelligence Generation), a novel, embedding-free paradigm for simulating the output of KGEs that uses a tiny fraction of the parameters. TWIG learns weights from inputs that consist of topological features of the graph data, with no coding for latent representations of entities or edges. Our experiments on the UMLS dataset show that a single TWIG neural network can predict the results of state-of-the-art ComplEx-N3 KGE model nearly exactly on across all hyperparameter configurations. To do this it uses a total of 2590 learnable parameters, but accurately predicts the results of 1215 different hyperparameter combinations with a combined cost of 29,322,000 parameters. Based on these results, we make two claims: 1) that KGEs do not learn latent semantics, but only latent representations of structural patterns; 2) that hyperparameter choice in KGEs is a deterministic function of the KGE model and graph structure. We further hypothesise that, as TWIG can simulate KGEs without embeddings, that node and edge embeddings are not needed to learn to accurately predict new facts in KGs. Finally, we formulate all of our findings under the umbrella of the ``Structural Generalisation Hypothesis", which suggests that ``twiggy" embedding-free / data-structure-based learning methods can allow a single neural network to simulate KGE performance, and perhaps solve the Link Prediction task, across many KGs from diverse domains and with different semantics.
\end{abstract}


\section{Introduction}
Knowledge Graphs (KGs) are graph-based data stores in which all data is expressed as a series of labelled nodes connected by a set of labelled edges; every set of a ``subject" (or head) node $s$, a ``predicate" edge $p$, and an ``object" (or tail) node $o$ is called an $(s,p,o)$ triple \cite{kg-ovewview}. The ability to represent data as triples in a graphical form has become particularly useful in the domains of bioinformatics and computational biology, where massive biological and biomedical datasets can be directly and intuitively represented as a network of relations and interactions \cite{bio2rdf,umls,biokg,ckg,drkg,ogb,PharmKG,OpenBioLink}. However, despite the power of this graph-based format, much of the data far exceeds the size that can be easily analysed by humans without the use of computational tools. The domain of Knowledge Graph Embeddings aims to address this gap by producing low-dimension embeddings of nodes and edges in a graph that can be used to both summarise the graph data and predict new links \cite{kge-survey,rml-review,kge-completion-rev,kg-ovewview}.

Specifically, all KGE models are trained to solve the Link Prediction (LP) Task, which consists of answering queries in the form $(s,p,?)$ or $(?,p,o)$, where $?$ represents the subject or object to be predicted \cite{kge-survey,rml-review,kge-completion-rev,kg-ovewview}. It does this by learning embeddings for each node and edge that can be used to assign a plausibility score to every statement $(s,p,o)$ that should be high if $(s,p,o)$ occurs in the KG, and low if it does not \cite{light-into-the-dark,kge-survey,rml-review,kge-completion-rev,kg-ovewview}. In other words, KGEs are trained to summarise the information contained in a KG in low-dimensional latent space in a way that distinguishes true facts from false statements \cite{kge-survey,rml-review,kge-completion-rev,kg-ovewview}.

The literature around KGE models generally attributes ``latent semantics" to the learned embeddings -- it assumes that the learned features represent a higher level of conceptual knowledge about the graph \cite{kge-survey,rml-review,kge-completion-rev}. The underlying assumption in this presentation of KGE methods, which has remained largely unquestioned, is the assumption that KGEs have learned a higher-order semantic representation of the knowledge graph that necessarily extends beyond what could be learned by simply memorising common graph structures to replicate \cite{kge-survey,rml-review,kge-completion-rev}. 

Similarly, with the partial exception of some recent works targeted at specific parts of KGE model construction (see \cite{neg-samp-analysis,loss-func-analysis,old-dog-new-tricks}), established KGE literature assumes that hyperparameter searching is necessary to determine at least some optimal model settings \cite{light-into-the-dark}. Moreover, there is no established hypothesis that optimal hyperparameters can be predicted in a pre-hoc manner before running a full search \cite{kge-survey,rml-review,kge-completion-rev,kg-ovewview}. This means, that, to date, essentially all hyperparameter selection is performed using some manner of hyperparameter search over a large range of possible combinations.

We summarise the above as three core assumptions that existing KGE literature makes:
\begin{enumerate}
  \item \textbf{The Embedding Assumption}: that embeddings are necessary to solve, or at least best suited to solving, the Link Prediction task,
  \item \textbf{The Latent Semantics Assumption}: that KGE models learn embedding semantics representing higher order conceptual knowledge, which they then apply to make predictions on KG data,
  \item \textbf{The Hyperparameter Stochasticity Assumption}: that hyperparameters must be explicitly searched for, and particularly that optimal hyperparameters cannot be predicted in a pre-hoc manner.
\end{enumerate}

We highlight that no study known to the authors, save \cite{hp-struct-basis} alone, has specifically questioned the validity of these assumptions on modern KGE models and KG learning approaches. Reference \cite{hp-struct-basis} is notable for explicitly questioning Hyperparameter Stochasticity Assumption, and hypothesising that optimal hyperparameters are a deterministic function of global KG structure.

In this work, we produce a new analytic paradigm to analyse all three of these core assumptions in the context of KG learning. The core contribution of our work is a novel Neural Network (NN) called TWIG (Topologically-Weighted Intelligence Generation) that is able to:
\begin{enumerate}
  \item simulate the output of KGE models in terms of both the ranked list predictions and the overall predictive performance,
  \item perform this prediction using only graph structure and hyperparameter settings as input features; notably using no node or edge embeddings of any form,
  \item perform these predictions over the entire grid of searched hyperparameters with a single TWIG model, thereby achieving dramatic reduction in parameter use and computational and memory cost.
\end{enumerate}

We perform all of our experiments on the state-of-the-art ComplEx-N3 model \cite{complex-n3} and the biological UMLS dataset \cite{umls}, a dataset that despite its small size sees consistent and important use in the biomedical domain \cite{light-into-the-dark}. While our analysis at the moment is limited by the use of a single model and dataset for our analysis, our novel TWIG model shows remarkably strong results in simulating the output of KGEs across all hyperparameter settings. We therefore postulate that further work in this area will be of immediate and direct benefit. We expect that such further work will deepen our understanding of the mechanisms by which KGEs work, expand our knowledge of KG structure-hyperparameter relations, and enable dramatic reductions in parameter use.

Furthermore, the existence of a high-performance neural network such as TWIG that can replicate the results of KGEs while only using global features consisting of hyperparameter configurations and graph structure provides strong initial evidence of three core hypotheses, which are the major theoretical results of this work. These hypotheses are as follows: 
\begin{enumerate}
  \item The Structural Learning Hypothesis: That KGEs do not learn latent semantics in embeddings, but rather only learn to implicitly summarize graph structure,
  \item The Hyperparameter Determinism Hypothesis: That optimal hyperparameter choice is a deterministic function of graph structure and the KGE model being used.
  \item The TWIG Hypothesis: That node / relationship embeddings are not necessarily needed to solve the Link Prediction (LP) task, but instead that learnable parameters could all be part of a single NN.
\end{enumerate}

We are able to provide initial evidence for all of these hypotheses for the ComplEx-N3 model and the UMLS dataset, and we propose that they will hold across other KGE models and KG datasets. We note, however, that while TWIG can simulate the output of KGEs, it is not currently able to solve the Link Prediction Task; as a result, there is less direct evidence for the TWIG Hypothesis as compared to the Structural Learning and Hyperparameter Determinism Hypotheses. Further exploration of these hypotheses is left to future work.

We further use the above hypotheses, and the graph-agnostic properties of TWIG, to propose \textbf{The Structural Generalisation Hypothesis}. This hypothesis states that since simulated learning methods such as TWIG allow prediction of the outputs of KGEs without using embeddings, that these simulated learning methods can be extended to allow generalised learning of graphs across KGs. While we are unable to fully test and evaluate this hypothesis in this work, existing literature taken in tandem with our empirical results provides strong initial support at the theoretical level.

The remainder of this paper is structured as follows. Section 2 provides an overview of related works in the literature. Section 3 provides basic preliminaries and formal hypothesis formulations needed to robustly present our methods and results. Section 4 presents the methods for all of our experiments, and forms the bulk of our paper. Section 5 presents our results and discusses them in the context of the existing state-of-the-art. Finally, Section 6 concludes the paper, discusses the limitations of this work, and outlines the major future directions that remain open.

All code and datasets needed to reproduce the experiments in this paper are included at the link \url{https://github.com/Jeffrey-Sardina/TWIG-release-1.0.git}.

\section{Related Works}
To the extent of the knowledge of the authors, no techniques currently exist that allow embedding-free learning of Knowledge Graphs nor embedding-free simulation of KGE models. Knowledge Graph Embeddings are by definition embedding-based \cite{rml-review,kge-survey,kge-completion-rev,kg-ovewview}; existing GNN approaches such as GCN, GraphSage, CensNet, IDGL, and many others all create embedding vectors for nodes / edges \cite{censnet,idgl,gnn-overview}. Similarly, only one work known to the authors (\cite{hp-struct-basis}) has attempted to explicitly and methodologically question the Hyperparameter Stochasticity Assumption, suggesting that hyperparameter choice is in fact a deterministic function of graph structure. No work known to the authors questions either the Embedding Assumption or the Latent Semantics Assumption.

Some work has been done in creating sub-node embeddings for KGEs, which results in a significant reduction of memory needed to run KGEs. This is exemplified in the recent NodePiece model, which uses anchor nodes and edges to tokenise embeddings for all other nodes and edges \cite{nodepiece}. However, NodePiece remains an embedding-based method, with explicit embeddings for all anchors and implicit (tokenised) embeddings for all non-anchor nodes and edges \cite{nodepiece}.

There has been some previous work in determining the relationship between optimal hyperparameters, KGE model choice, and graph structure \cite{light-into-the-dark,old-dog-new-tricks,neg-samp-analysis,loss-func-analysis,kges-for-lp-compare}. In particular, \cite{neg-samp-analysis} analysed optimal negative sampler choice for KGE models in terms of KGE model choice and graph structure. Reference \cite{loss-func-analysis} found similar relations for optimal loss function choice, and \cite{kges-for-lp-compare} analysed optimal KGE model choice for a given dataset. Two substantially broader studies \cite{light-into-the-dark,old-dog-new-tricks} documented overall KGE model performance on a large variety of datasets; however, neither of these specifically analysed the impact of individual hyperparameters as a function of graph structure.

\section{Preliminaries}
\subsection{Literature Definitions}
\textbf{Knowledge Graphs.} A Knowledge Graph, $G = \{E, R\}$ is set of entities $E$ and a set of directed, labelled relations $R$ \cite{kg-ovewview}. Each statement in a Knowledge Graph is the 3-element triple consisting of a subject, a predicate, and an object; this is written $(s, p, o)$, where $s \in E, p \in R, o \in E$.

\textbf{Knowledge Graph Embedding.} Knowledge Graph Embedding (KGE) models convert a KG's nodes and edges to vector embeddings \cite{rml-review,kge-survey,kge-completion-rev,kg-ovewview}. Formally, this process can be defined as $(s, p, o) \rightarrow (e_s, e_p. e_o)$ where $s$, $p$, and $o$ represent the subject, predicate, and object in a triples and $e_s$, $e_p$, and $e_o$ their respective embeddings; $g$ and $e_g$ are defined analogously in the case that named graphs are embedded.

\textbf{Link Prediction Task.} The Link Prediction (LP) Task is to answer queries $(?, p, o)$ and/or $(s, p, ?)$ where $s \in E, p \in R, o \in E$ and $?$ is the entity being predicted to complete the query triple \cite{rml-review,kge-survey,kge-completion-rev,kg-ovewview}. KGE models solve this by learning to maximise the difference between the scores of `positive' triples observed in training and randomly generated (`negative') triples not in the training dataset, according to a model-specific scoring function \cite{kge-survey,rml-review,kge-completion-rev,neg-samp-analysis}.

\textbf{KGE model.} A KGE model is an embedding-based model that solves the Link Prediction task. The output of these models is a rank-ordered list $l_E$ of all nodes $E$ that could answer the LP query $(?, p, o)$ or $(s, p, ?)$. $l_E$ is ordered such that nodes with the lowest rank (i.e. the rank closest to 1) are considered the best answers, and nodes with higher rank are considered less probable answers. As such, KGE models can be described as embedding based models that learn to rank.

\textbf{Mean Reciprocal Rank.} Given a ranked list of true triples versus all possible generated negatives (formed by randomly corrupting the subject or object) the Mean Reciprocal Rank (MRR) metric takes the mean of the reciprocal of all the ranks of the known true triples as a performance score. It is necessarily bounded on (0, 1], with higher values indicating better performance.

\subsection{Problem Specification}
KGE models learn to minimise the rank of known-true triples and maximise that of all other triples \cite{kge-survey,rml-review,kge-completion-rev,neg-samp-analysis}. The result of this is graph-specific embeddings -- the embedding found using one KGE model for one dataset cannot be applied directly to any other dataset \cite{kge-survey,rml-review,kge-completion-rev,kg-ovewview}.

While this approach has been successful -- being the universal foundation for all KGE methods known to the authors to-date \cite{kge-survey,rml-review,kge-completion-rev,kg-ovewview,light-into-the-dark,old-dog-new-tricks} -- we propose an embedding-free learning paradigm that simulates the output of the link prediction task by learning to predict all ranks output by an existing KGE method. In doing this, we restrict our inputs to two types: local structural features of the query triple being predicted, and hyperparameters used by the original KGE model to learn that triple.

As such, this approach necessarily is about replicating KGE model behaviour using only structure and model settings, not latent features or any sort of graph semantics or logic. In other words, this is a \textit{simulation} procedure, and we refer to it as ``Learning to Simulate". Similarly, we refer to the task of learning to simulate KGEs as the ``KGE Simulation Task", to distinguish it from the Link Prediction Task.

We note that successfully solving this task would provide strong initial evidence for our three core hypotheses; i.e. the Structural Learning Hypothesis, the Hyperparameter Determinism Hypothesis, and the TWIG Hypothesis.

\section{Methods}
\subsection{Dataset Choice}
For our study, we chose UMLS, a standard biomedical KG commonly used both in biological applications and as a benchmark dataset for KGEs \cite{umls,light-into-the-dark}. In particular, we chose UMLS over the more standard FB15K-237 and WN18RR KGs for two reasons. First, it is directly applicable to the biomedical domain. Second, it is substantially smaller than either FB15K-237 or WN18RR, which allowed us to perform multiple replicated experiments on a large hyperparameter grid, something that would be computationally difficult on the other datasets even using powerful modern computational hardware \cite{umls,light-into-the-dark}. 

The overall structure of the UMLS dataset is given in Table \ref{tab:umls}. Note that degree refers to total degree; i.e. the number of edges incident on a particular node, regardless of whether the relationship is an incoming edge or an outgoing edge.

\begin{table}
    \centering
    \begin{tabular}{l|l}
        \textbf{ Structural Feature}&\textbf{Value}\\ \hline
         Num. nodes& 135\\
         Num. relations& 46\\
         Num. triples& 6,529\\
         Min. degree& 4\\
         Median degree& 71\\
         Max. degree& 382\\
    \end{tabular}
    \caption{Overall structure features of the UMLS dataset}
    \label{tab:umls}
\end{table}

\subsection{Model Space Search}
Given that ComplEx-N3 (the ComplEx KGE model using the $L_3$ regulariser) is established as the de-facto state-of-the-art KGE model for many applications \cite{complex-n3}, as well as remaining one of the strongest overall KGE models \cite{baselines-strike-back-2,light-into-the-dark}, we performed all of our experiments on the ComplEx-N3 model. We trained different ComplEx-N3 models under a broad grid of hyperparameter values to determine relative performance in each case; the grid used is shown in Table \ref{tab:hp-grid}. A technical description of all of these hyperparamters and their definitions can be found in \cite{light-into-the-dark}.

All training was done on the training set, and evaluations of hyperparameter performance were done on the validation set. All experiments were run using the Adam optimiser for 100 epochs. 

\begin{table}
    \centering
    \begin{tabular}{p{3cm}|p{5cm}}
         \textbf{Hyperparameter}& \textbf{Values Searched}\\ \hline
         Negative Sampler& Basic, Bernoulli, Pseudo-Typed\\
        \#Negatives per Positive& 5, 25, 125\\
         Loss Function& Margin Ranking, Binary Cross Entropy (wth Logits), Cross Entropy\\
         Margin (if applicable)& 0.5, 1, 2\\
         Learning Rate& 1e-2, 1e-4, 1e-6\\
         Embedding dimension& 50, 100, 250\\
         Regularisation Coefficient& 1e-2, 1e-4, 1e-6\\
    \end{tabular}
    \caption{The grid of hyperparameters used in the experiments}
    \label{tab:hp-grid}
\end{table}

For all hyperparameter settings, we recorded two outputs: the Mean Reciprocal Rank (MRR) score of the KGE model overall, as well as the ranked-list predictions for every triple in the validation set against its corruptions.

We repeated this procedure 4 times, to produce 4 sets of size-1215 grids of results. One of these grids was separated as a hold-out test-set; the remaining 3 were used as training sets, as described below.

\subsection{Data Modelling}
\subsubsection{Feature Selection}
The result of running the model space search was a grid of 1215 distinct hyperparameter combinations, as well as the individual ranked-list results and Mean Reciprocal Rank (MRR) scores for all those combinations. As our goal was to simulate the task of Knowledge Graph embedding, we framed our problem of constructing TWIG as predicting the rank that would be assigned to each individual triple in the UMLS validation set used to evaluate the KGEs.

In order to provide a valid test of our hypothesis, TWIG must take the form of a function $f(structure, hyperparameters) \rightarrow ranked list$. We present this as a regression problem where the hyperparameter features are the values of the hyperparamters used for each experiments. For structure, we represented this as local structure about the triple whose score is being predicted. Noting that existing KGE literature demonstrates that KGEs only learn how to make predictions for a triple from the 1 or 2-hop distance around that triple \cite{topological-imbalance,PoLo,kge-poisoning,kge-poisoning-2,gradient-rollback}, we limit the local structural features that are included to those that can be mined in the local 2-hop distance about the triple being predicted. Note that when computing structural values, we compute them \textit{only as seen in the train set of the UMLS KG} so that there is no data leakage from the graph's validation or test sets.

Overall, the full set of structural and hyperparameter features included are given in Table \ref{tab:fts}. We note that the validation set used in our evaluation contains 1304 queries in total.

\begin{table}
    \centering
    \begin{tabular}{p{3cm}|p{5cm}}
         \textbf{Feature}&\textbf{Meaning}\\ \hline
 \textit{Hyperparameter Features}&\\
         Negative Sampler
&The negative sampling strategy used\\
         \#Negatives per Positive
&The number of negatives samples for each positive triples during training\\
         Loss Function
&The loss function used\\
         Margin (if applicable)
&The margin used in the loss function (if applicable)\\
         Learning Rate
&The learning rate for the Adam optimiser\\
         Embedding dimension
&The dimension of KGE model embeddings\\
         Regularisation Coefficient&The coefficient multiplies to the regulariser\\
          &\\
         \textit{Structural Features}&\\
 is\_head&Whether the part of the triple being corrupted is the head (i.e. (?,p,o)) or the tail (i.e. (s,p,?))\\
 s\_deg&The degree of the subject node in the training set\\
 o\_deg&The degree of the object node in the training set\\
 p\_freq&

The frequency of the predicate in the training set\\
 s\_p\_cofreq&The number of times the given subject and predicate co-occur in the training set\\
 o\_p\_cofreq&The number of times the given subject and predicate co-occur in the training set \\
 s\_o\_cofreq&The number of times the given subject and object co-occur in the training set\\
 s min deg neighbnour&The degree of the lowest-degree neighbour of the subject node in the training set\\
 s max deg neighbnour&The degree of the highest-degree neighbour of the subject node in the training set\\
 s mean deg neighbnour&The degree of the mean-degree neighbour of the subject node in the training set\\
 o min deg neighbnour&The degree of the lowest-degree neighbour of the object node in the training set\\
 o max deg neighbnour&The degree of the highest-degree neighbour of the object node in the training set\\
 o mean deg neighbnour&The degree of the mean-degree neighbour of the object node in the training set\\
 s num neighbnours&the total number of neighbours the subject node has in the training set\\
 o num neighbnours&the total number of neighbours the object node has in the training set\\
 s min freq rel&The frequency of the lowest-degree neighbour of the subject node in the training set\\
 s max freq rel&The frequency of the most-frequent-occurring edge incident on the subject node in the training set\\
 s min freq rel&The mean frequency of edges incident on the subject node in the training set\\
 o min freq rel&The frequency of the lowest-degree neighbour of the object node in the training set\\
 o min freq rel&The frequency of the most-frequent-occurring edge incident on the object node in the training set\\
 o min freq rel&The mean frequency of edges incident on the object node in the training set\\
 s num rels&The total number of relations incident on the subject node in the training set\\
 o num rels&The total number of relations incident on the object node in the training set\\
    \end{tabular}
    \caption{A summary of all input features used, and their definitions, in the TWIG model}
    \label{tab:fts}
\end{table}

\subsubsection{Determination of Signal}
Once our features were prepared, we then performed an analysis of what sources of signal we could use to learn on this data and simulate KGE using a single neural network. In particular, we found two critical sources of signal that could be very readily learned, and very easily distinguished from noise. These were:
\begin{enumerate}
  \item near-1 correlation of MRRs from KGE models run on the same hyperparameters but with different random seeds
  \item near-0 KL Divergence of the distribution of values in output ranked lists from KGE models run on the same hyperparameters but with different random seeds
\end{enumerate}

We will discuss both of these choices in turn. With respect to mining the high correlation of MRR across experiments run on the same hyperparameters but with different random seeds, we observed that the correlation of MRR values between our four rounds of hyperparameter validation were greater than 0.99 in all cases. This is shown in Table \ref{tab:mrr-corr}. Since correlation itself is not a loss function, we instead follow the protocol common in most regression systems and used Mean Squared Error (MSE) loss between the predicted and true MRR values as a proxy metric to mine the extremely high correlation between MRRs. 

\begin{figure}
    \centering
    \includegraphics[width=1\linewidth]{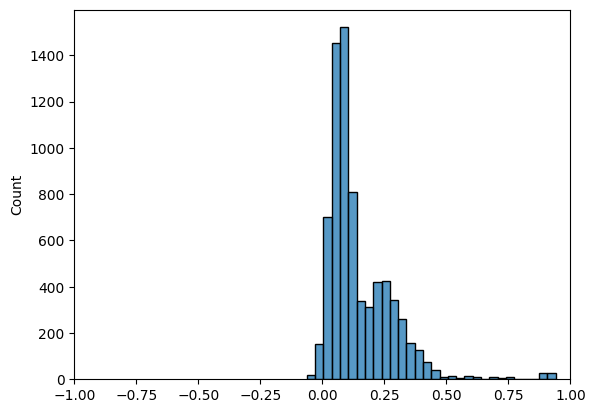}
    \caption{Distribution of the Pearson correlation values between all ranked lists using the same hyperparameter configurations.}
    \label{fig:rl-corr}
\end{figure}

We could not use correlation between ranked lists as a source of signal, however, because the correlation between ranked lists was exceedingly low, often near 0, indicating that it would not be useful as signal during learning. Figure \ref{fig:rl-corr} shows the distribution of Pearson-correlated values for all combinations of ranked lists that used the same hyperparameters but different random seeds. Due to the tendency of these to be centred near zero, we concluded that they were not a reliable source of signal for learning.

We note that the KL Divergence of the distribution of values in output ranked lists from KGE models run on the different hyperparameters was substantially higher than those run on the same hyperparameter under different initialisations; see Table \ref{tab:kl-div} for average KL Divergence values between experiments with all matching hyperparamters (calculated as the KL Divergence between histograms with 30 bins of the same width formed from the values in the ranked list). In contrast to this, taking the average KL Divergence using the same methods of all non-identical sets of hyperparameters, the average KL Divergence value was $0.3632$ -- substantially higher than those obtained under the same hyperparameters. This indicates that KL Divergence between ranked lists from experiments with the same hyperparamters is much lower than that expected between experiments with different hyperparameters, suggesting it is a strong source of signal for learning. Since KL Divergence itself is a loss function, we used KL Divergence between the distributions of the values in the predicted and true ranked lists as our second loss function during learning.

During training, we used the same manner of calculation for the distribution of ranks in the ranked list: the KL Divergence between histograms with 30 bins of the same width formed from the values in the predicted and true ranked lists. However, since the exact counting-based binning operation is not differentiable, we replaced it with a "soft histogram" version that assigns values into bins using the differentiable Sigmoid function rather than the non-differentiable binary step function. A reference implementation of this function is provided with the rest of our code in our published codebase.

Finally, we found that both of these sources of signal (MSE and KL Divergence) require a full ranked list, not a single output value, to be computed. In other words, all elements of the ranked list have to be predicted before loss is calculated. As a result, even though our input features are vectors, we can only calculate loss at the level of all aggregated predictions. To do this, we construct batches for each unique hyperparameter configuration (of 1215 total) and random seed (of 3 iterations total in the training set), resulting in 3645 batches in total. All predictions for a batch are then aggregated, used to compute the distribution of ranks and MRR, and then passed into our loss functions during training.

\begin{table}
    \centering
    \begin{tabular}{ccccl}
         &  Exp run 1&  Exp run 2& Exp run 3&Exp run 4\\
         Exp run 1&  1&  &  &\\
         Exp run 2&  0.994&  1&  &\\
         Exp run 3&  0.994&  0.995&  1&\\
         Exp run 4& 0.9939& 0.9943& 0.9951&1\\
    \end{tabular}
    \caption{Pairwise correlations of the MRR scores for all 1215 hyperparameter combinations across 4 runs using different random seeds for for each run and hyperparameter combination}
    \label{tab:mrr-corr}
\end{table}

\begin{table}
    \centering
    \begin{tabular}{ccccc}
         &  Exp run 1&  Exp run 2&  Exp run 3& Exp run 4
\\
         Exp run 1&  0&  &  & 
\\
         Exp run 2&  0.0250&  0&  & 
\\
         Exp run 3&  0.0250&  0.0245&  0& 
\\
         Exp run 4&  0.0248&  0.0248&  0.0247& 0\\
    \end{tabular}
    \caption{Average KL Divergence values of the distributions of values in output ranked lists  for all 1215 hyperparameter combinations across 4 runs using different random seeds for for each run}
    \label{tab:kl-div}
\end{table}

\subsubsection{Interpretation of our Modelling Approach}
Our choice to model the atomic unit of learning as a batch of inputs representing all triples being predicted has a few important theoretical properties. All parameters are updated based not on how a specific rank was predicted, but by how the list of ranks as a whole was predicted as a broader scale. We note that none of our loss terms enforce explicit order on this ranked list. This means that which triple is assigned which rank is left free under the condition that, at a global scale, the order-less set of ranks assigned to the set of triples still matches the expected distribution and has the expected MRR.

Another crucial property of the rank lists themselves is that ranked lists run on the same hyperparamters, but with different random seeds, correlate very poorly (see Figure \ref{fig:rl-corr}). This indicates that different parts of a graph are learned differently based only on random initialisations, and that the same fact could be learned well or not purely due to chance. However, the near-1 correlation of MRRs with identical hyperparameters but across different random seeds (see Table \ref{tab:mrr-corr}) indicates that overall learning is highly robust in th face of random initialisations. This effect is theoretically critical both for KG learning in general and for understanding the operation of KGE models in deployment. It means that any attempt to enforce an explicit hard order on ranked-list output would likely decrease performance. Instead, either a soft / partial order, or no explicit loss term to enforce ordering, are best suited here. While we have taken the no-explicit-ordering approach here, we leave analysis of a soft, partial-order based system to future work.

\subsection{Constructing TWIG}
TWIG was constructed as a neural network made of exclusively Dense Layers that form three key components. After taking input (a vector of features describing hyperparameters and KG structure around the triple being predicted, as given in Table \ref{tab:fts}) it splits this input into two parts: one describing the hyperparameters and one describing the structural elements. Both of these input feature sub-vectors are then passed through their respective learning components to produce hidden representations of the input data. These are the Hyperparameter Learning Component (on the left, in green) and the Structure Learning Component (on the right, in blue), respectively. The output of each of these components is then concatenated and passed into the third and final Integration Component (at the bottom, in black) that produces a hidden representation integrating the information learned from both the structural data and the hyperparameter data. The output of TWIG is a single value for each input vector, which represents the rank assigned to the triples query represented by the input. Note that the output is passed through the ReLU activation function and incremented by 1 because the output, which represents a predicted rank, is defined as a strictly positive value greater than or equal to 1.  

\begin{figure}
    \centering
    \includegraphics[width=1\linewidth]{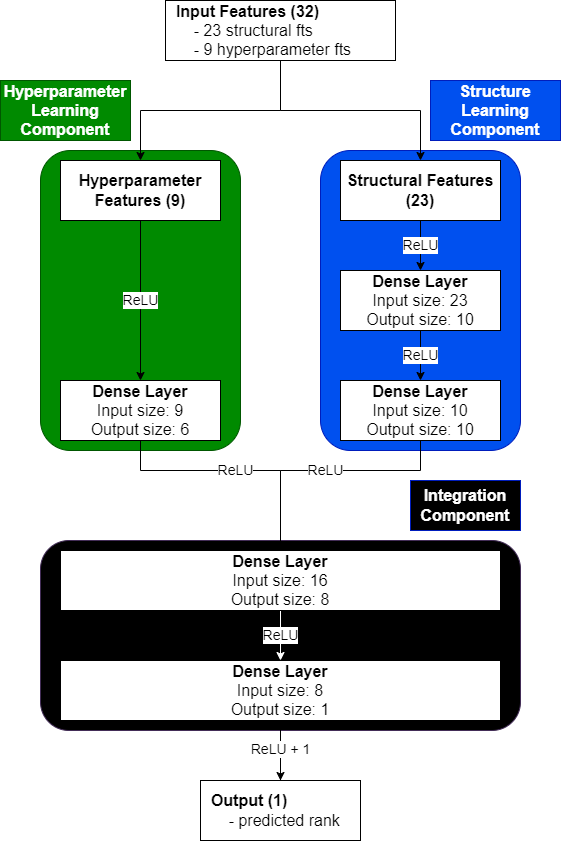}
    \caption{The TWIG neural architecture, based off of three core components for learning hyperparameter representations (green), structure representations (blue) and for integrating those representations to produce output (black).}
    \label{fig:twig-nn}
\end{figure}

Figure \ref{fig:twig-nn} shows the neural architecture of TWIG in detail, including the number of Dense Layers in each component and the input and output sizes of each. ReLU is used as the activation function between all Dense Layers in the network.

For training, we adopted a 2-tiered approach. We trained TWIG for 50 epochs using only KL Divergence between ranked lists as a loss function. After this, we froze the parameters in all layers save the final layer, and trained for another 100 epochs using both KL Divergence between ranked lists and Mean Squared Error (MSE) loss between predicted and true MRR scores as loss functions. As such, a total of 150 epochs were used for training.

\section{Results and Discussion}
Our core result is this: \textit{The fully-trained TWIG model achieves an R2 value of 86.18\% for predicting MRR of a hyperparameter combination, and does this by predicting the entire ranked list using only localised graph features and hyperparameter setting information}. This R2 value means that TWIG is able to explain 86.18\% of the variation in MRR seen across all different hyperparameter settings using KG structure and model configuration only as input, and corresponds to an overall correlation of 0.9371. While there are no existing baselines known to the authors to which we can compare, we can conclude that this outperforms the simplest average-MRR baseline, as such a function is constant and therefore has an R2 value of 0\%. These results in themselves have huge implications, which we will address in turn: the Structural Learning Hypothesis, the Hyperparameter Determinism Hypothesis, and the TWIG Hypothesis, 

\textbf{The Structural Learning Hypothesis} states that KGEs do not learn latent semantics in embeddings, but rather only learn to implicitly summarize graph structure. Our results here, showing that TWIG can simulate the output of ComplEx-N3 across hyperparameter settings, prove that using structure only, without any latent semantics of the KG, is enough to replicate almost entirely the predictions of KGEs run on the UMLS dataset. This indicates that KGEs on the UMLS dataset are not performing any significant learning that cannot be summarised by structure. We hypothesise that the Structural Learning Hypothesis will hold across other KGE models and datasets, but leave testing that hypothesis for future work.

\textbf{The Hyperparameter Determinism Hypothesis} states that optimal hyperparameter choice is a deterministic function of graph structure and the KGE model being used. Once again, we show that this hypothesis holds for the ComplEx-N3 KGE model and the UMLS dataset -- the extremely high correlation and R2 value seen indicate that the TWIG NN is such a function that can deterministically map hyperparameters to model performance. This means that, for the model and dataset tested, optimal (and indeed all) hyperparameters are necessarily a deterministic function of graph structure. We hypothesise that the Hyperparameter Determinism Hypothesis will hold across other KGE models and datasets, but leave testing that hypothesis for future work.

\textbf{The TWIG Hypothesis} states that node / relationship embeddings are not needed to solve the LP task. Since TWIG is able to construct a ranked list from localised graph structure, and since this ranked list has an MRR that matches with very high precision that of the ground-truth MRR, our data suggests that embeddings are not needed to simulate the output of ComplEx-N3 trained on UMLS, and further suggests that this simulation could feasibly extend to other KGE models and datasets. However, whether embedding-free (``twiggy'') methods can solve the Link Prediction Task in full, rather than just simulate the output of KGEs, remains to be tested. We hypothesise that twiggy, embedding-free LP is possible on basis of the success of our TWIG model in simulating KGEs. We further hypothesise that this will hold across KGE models and datasets, but leave testing this as a future direction. 

Overall, these results suggest that learning is driven not by latent semantics, but by summarisation of data structure. All 1215 experiments in the test set had a combined parameter cost of 29,322,000 parameters. TWIG, which uses only 2,590 parameters, recapitulated the results of all of those experiments -- in other words, it simulated their ranked list and MRR output with high accuracy while using 0.008832\% of the parameters needed to originally produce the data. Thus, TWIG represents the outputs of KGE models in a highly compact manner.

Finally, we propose \textbf{The Structural Generalisation Hypothesis.} Given that that TWIG can lead to massive reductions in parameter use, and supposing that the above three hypotheses hold, we propose that TWIG can generalise not only over different hyperparameter settings, but also over different Knowledge Graph datasets. Since all KGs can be annotated by the same structural features, and since KG-specific embeddings are no longer needed, TWIG would in theory allow ranked-list and MRR computation on novel datasets from their structure alone. This suggests that TWIG could lead to a massive step forwards in the generalisability of graph models and in graph learning. Moreover, if specific embeddings neither contain higher-order semantics (the Structural Learning Hypothesis), then this method should be expected to match the performance of existing state-of-the-art KGE models, without the need for hyperparameter searching or KG-specific learning. We note that existing literature has already demonstrated that KGEs learn different regions of graph differently based on degree \cite{topological-imbalance}, suggesting that the Structural Generalisation Hypothesis has a theoretical basis in existing literature as well.

\section{Conclusion and Future Work}
In this work, we propose three novel hypothesis: the Structural Learning Hypothesis, the Hyperparameter Determinism Hypothesis, and the TWIG Hypothesis. The core idea behind all three of these hypothesis is that latent semantics are neither needed, nor actually learned, in KGEs -- that learning and hyperparameter choice are instead driven entirely by graph structure. Our new TWIG model is able to simulate the output of KGE models across 1215 hyperparameter configuration with high fidelity (R2 = 0.8618\%) while using only 0.008832\% of the parameters.

Our work here is limited primarily by the scope of data available -- our analysis was limited to the state-of-the-art ComplEx-N3 KGE model and the biomedical dataset UMLS. More work is needed to assess the validity of our findings on other KGE models and other datasets. However, non-withstanding these notable limitations, our work is the first of its kind known to the authors to take a ground-up, structure-based analysis of KGs and KGE models, to suggest hyperparameter determinism rather than stochasticity, and to provide evidence that embeddings may not be needed to solve the Link Prediction Task. The authors intend to follow up on this work with much more expansive studies to assess to what extent these findings are general to KG learning.

\section*{Acknowledgment}
The authors would like to thank Alok Debnath (ORCID 0000-0002-1270-369X) for his critical insights into the use of the Sigmoid function for the construction of a differentiable ``soft histogram" used as a core component of TWIG, as well as for his feedback on the manuscript.

This research was conducted with the financial support of Science Foundation Ireland D-REAL CRT under Grant Agreement No. 18/CRT6225 at the ADAPT SFI Research Centre at Trinity College Dublin, together with sponsorship of Sonas Innovation Ireland. The ADAPT SFI Centre for Digital Content Technology is funded by Science Foundation Ireland through the SFI Research Centres Programme and is co-funded under the European Regional Development Fund (ERDF) through Grant \# 13/RC/2106\_P2.

\bibliographystyle{splncs04}
\bibliography{TWIG}

\end{document}